\documentclass[10pt,journal,compsoc]{IEEEtran}
\ifCLASSOPTIONcompsoc
  \usepackage[nocompress]{cite}
\else
  \usepackage{cite}
\fi
\ifCLASSINFOpdf
\else
\fi
\usepackage{amsfonts}
\usepackage{amsmath}
\newtheorem{theorem}{Theorem}
\newtheorem{lemma}{Lemma}
\newtheorem{corollary}{Corollary}

\usepackage{booktabs}
\usepackage{multirow}
\usepackage{graphicx}
\usepackage{tabularx}
\usepackage{array}
\usepackage{url}

\usepackage{enumerate}

\begin{document}
%
\title{Block-regularized $5\times 2$ Cross-validated McNemar's Test for
Comparing Two Classification Algorithms}
%
%
%
%

\author{Jing~Yang, Ruibo~Wang, Yijun~Song, Jihong~Li
\IEEEcompsocitemizethanks{\IEEEcompsocthanksitem J. Yang is with the School of 
Automation and Software Engineering, Shanxi University, Taiyuan, China, 030031.

R. Wang, Y. Song, and J. Li are with the 
School of Modern Educational Technology, Shanxi University, Taiyuan, China, 
030006.\protect\\
E-mail: \{yangjing, wangruibo, songyj, li\_ml\}@sxu.edu.cn}
\thanks{Manuscript received April 19, 2005; revised August 26, 2015.} 
}

%
%

\markboth{Journal of \LaTeX\ Class Files,~Vol.~14, No.~8, August~2015}%
{Wang \MakeLowercase{\textit{et al.}}: Block-Regularized $5\times 2$ 
Cross-Validated McNemar's Test}
%



\IEEEtitleabstractindextext{%
\begin{abstract}
In the task of comparing two classification algorithms, the widely-used
McNemar's test aims to infer the presence of a significant difference between
the error rates of the two classification algorithms. However, the power of
the conventional McNemar's test is usually unpromising because the hold-out (HO)
method in the test merely uses a single train-validation split that usually
produces a highly varied estimation of the error rates. In contrast, a
cross-validation (CV) method repeats the HO method in multiple times and
produces a stable estimation. Therefore, a CV method has a great advantage to
improve the power of the McNemar's test. Among all types of CV methods, a
block-regularized $5\times 2$ CV (BCV) has been showed in many previous studies
to be superior to the other CV methods in the comparison task of algorithms 
because the $5\times 2$ BCV can produce a high-quality estimator of the error 
rate by regularizing the numbers of overlapping records between all training 
sets. In this study, we compress the 10 correlated contingency
tables in the $5\times 2$ BCV to form an effective contingency table. Then,
we define a $5\times 2$ BCV McNemar's test on the basis of the effective 
contingency table. We demonstrate the reasonable type I error and the 
promising power of the proposed $5\times 2$ BCV McNemar's test on multiple 
simulation and real-world data sets.
\end{abstract}

}

\maketitle

\IEEEdisplaynontitleabstractindextext

%
\IEEEpeerreviewmaketitle

\IEEEraisesectionheading{\section{Introduction}\label{sec:introduction}}

\IEEEPARstart{I}{n} machine learning domain, the popular McNemar's test aims to
compare the error rates of two classification algorithms and choose the better
one. It has been widely applied in various real-world applications, including 
image classification \cite{RN1230,RN1229}, speech and emotion recognition 
\cite{RN714,Erdem_2010}, sentiment classification \cite{RN1226,RN1227}, and 
word segmentation task \cite{RN1228}.

The conventional McNemar's test is typically conducted with a hold-out (HO) 
validation that merely uses a single train-validation split on a data set. 
The train set is used to train the two classification algorithms in a 
comparison, and the validation set is used to induce a $2\times 2$ contingency 
table that summarizes the counts of the inconsistent predictions between the two 
algorithms. Because the HO estimator of the error rate of an algorithm is 
usually unstable \cite{RN758,RN391,RN380}, the power of the conventional 
McNemar's test is unpromising \cite{RN781}.

Compared with an HO validation, a cross-validation (CV) repeats HO validation
in multiple times and averages all HO estimators to produce a more stable 
estimator of the error rate \cite{RN112}. Therefore, a CV method has a great 
potential to improve the power of the McNemar's test. In this study, we aim to 
improve the McNemar's test with a CV method. Nevertheless, because all the HO 
validations in a CV method are performed on a single data set, the contingence 
tables and the McNemar's test statistics induced from the multiple HO 
validations in a CV are correlated. Therefore, the main challenge in the 
construction of a CV-based McNemar's test is how to aggregate the correlated
contingency tables to form a reasonable McNemar's test statistic.

An intuitive way to overcome the challenge is summing up all McNemar's test 
statistics in a CV method. Specifically, considering the immensely popular 
$K$-fold CV \cite{RN758,5342427}, the $K$ McNemar's test statistics in a 
$K$-fold CV are na\"{i}vely assumed to be independent and summed up to form a 
novel test statistic following a $\chi^2$ distribution with $K$ degrees of 
freedom (DoF). However, because there is a large overlap between the $K$ 
training sets in a $K$-fold CV when $K>2$, the $K$ McNemar's test statistics are 
correlated. Thus, the constructed test statistic are unreasonable. In this 
study, the intuitive McNemar's test is named a na\"ive $K$-fold CV McNemar's 
test, and it is not recommended. 

Previous studies illustrated that a block-regularized $5\times 2$ CV (BCV) has 
more advantages than the $K$-fold CV in the task of algorithm comparison 
\cite{RN524}. Specifically, Dietterich showed that the $5\times 2$ CV 
outperforms the $K$-fold CV in the paired $t$-tests for comparing classification
algorithms \cite{RN391}. Alpaydin further employed $5\times 2$ CV to develop a 
combined $F$-test \cite{RN722}. Furthermore, Wang et al. developed a 
$3\times 2$ BCV $t$-test and showed that it has comparable type I error and 
power with the $5\times 2$ CV paired $t$-test and $F$-test \cite{RN857}. Wang,
Li, and Li developed a calibrated $F$-test based on a $5\times 2$ BCV 
\cite{RN486}, and it achieves the state-of-the-art performance in the task of
algorithm comparison. In fact, a $5\times 2$ BCV regularizes the numbers of 
overlapping records between all training sets, and thus it enjoys a minimal
variance property and induces a stable estimator of the error rate of an
algorithm \cite{RN524}. Therefore, we consider using a $5\times 2$ BCV to 
develop a novel McNemar's test in this study.

Considering ten contingency tables in a $5\times 2$ BCV are correlated, 
instead of directly summing up the ten McNemar's test statistics, we compress
the ten correlated contingency tables into a virtual table named an effective 
contingency table through a reasonable Bayesian investigation where two
correlation coefficients between estimators of the disagreement probability in
the ten contingency tables are defined and analyzed. The effective contingency 
table helps us define a novel McNemar's test analogous to the conventional 
McNemar's test. Under a reasonable setting of the correlation coefficients, the
novel McNemar's test is named a $5\times 2$ BCV McNemar's test.

Extensive experiments are conducted on multiple synthetic data sets and
real-world data sets, involving several commonly used classification algorithms.
Type I error and power curve are used to measure the performance of a 
significance test for a comparison of two classification algorithms. Moreover,
14 existing significance tests are investigated in the experiments. The
experimental results illustrate that the proposed $5\times 2$ BCV McNemar's
test is superior to the other test methods. 

The rest of the paper is organized as follows. We investigate the conventional 
McNemar's test in Section \ref{sec:conv_mcn_test}. We then introduce a na\"{i}ve
$K$-fold CV McNemar's test in Section \ref{sec:naive_kfcv_mcn}. Our proposed 
$5\times 2$ BCV McNemar's test is elaborated in Section \ref{sec:5x2bcvmcn}. Our
experiments and observations are reported in Section \ref{sec:expr}, followed by
a conclusion in Section \ref{sec:conclusion}. Proofs of all theorems and lemmas 
are presented in Appendix.

\section{Conventional McNemar's Test}
\label{sec:conv_mcn_test}

\subsection{McNemar's test on an HO validation}

We let $D_n=\{z_i:z_i=(\boldsymbol{x}_i,y_i)\}_{i=1}^n$ be a data set consisting 
of $n$ iid records that are drawn from an unknown population $\mathfrak{D}$. In 
a record $z_i$, $\boldsymbol{x}_i$ is a predictor vector, and $y_i$ is the true 
class label of $\boldsymbol{x}_i$. We let $\mathcal{A}$ and $\mathcal{B}$ be the 
two classification algorithms that are compared on $D_n$. The comparison task is 
formalized into the following hypothesis testing problem:
\begin{equation}
H_0: \ \mu_{\mathcal{A}}-\mu_{\mathcal{B}}=0 \ \ vs. \ \  
H_1: \ \mu_{\mathcal{A}}-\mu_{\mathcal{B}} \neq 0, 
\label{eqn:core_problem}
\end{equation}
where $\mu_{\mathcal{A}}$ and $\mu_{\mathcal{B}}$ are the error rates of
$\mathcal{A}$ and $\mathcal{B}$, respectively, and they are also known as the 
generalization errors of $\mathcal{A}$ and $\mathcal{B}$ with the one-zero loss 
\cite{RN729}.

The problem in Eq. (\ref{eqn:core_problem}) can be addressed with a McNemar's 
test. The conventional McNemar's test is performed coupled with an HO 
validation. In an HO validation, $D_n$ is divided into two sub-blocks with a 
partition $(S,T)$ where $S$ and $T$ correspond to training and validation sets
such that $S\cup T = D_n$ and $S\cap T=\emptyset$. The sizes of training and
validation sets are $n_1=|S|$ and $n_2=|T|$, respectively.

On the basis of a partition $(S,T)$, algorithms $\mathcal{A}$ and $\mathcal{B}$
are trained on the training set $S$ and generate two classifiers, namely, 
$\mathcal{A}(S)$ and $\mathcal{B}(S)$. Then, the two classifiers are evaluated 
on the validation set $T$. The validation result can be summarized into a 
contingency table $\mathcal{C} = (n_{00}, n_{01},n_{10},n_{11})$ such that
$n_{00}+n_{01}+n_{10}+n_{11} =n_2$. Interpretation of $\mathcal{C}$ is given in 
Table \ref{tbl:interp_cont_table}. 

\begin{table}[t]
  \centering
  \caption{Interpretation of contingency table $\mathcal{C}$.}
  \label{tbl:interp_cont_table}
  \begin{tabular}{m{0.4cm}<{\centering}ccc}
  \toprule
  & & \multicolumn{2}{c}{Model $\mathcal{B}(S)$}\\
  & &Incorrect & Correct \\
  \midrule
  \specialrule{0em}{2pt}{2pt}
  \multirow{2}{*}{\rotatebox[origin=cB]{90}{Model $\mathcal{A}(S)$}} & 
  \rotatebox[origin=cB]{90}{Incorrect} & \multicolumn{1}{m{3cm}}{$n_{00}$: Count 
  of records in $T$ misclassified by both $\mathcal{A}$ and $\mathcal{B}$.} 
  & \multicolumn{1}{m{3cm}}{ $n_{01}$: Count of records in $T$ misclassified by 
  $\mathcal{A}$ but not by $\mathcal{B}$.} \\
  & \rotatebox[origin=cB]{90}{Correct}&\multicolumn{1}{m{3cm}}{ $n_{10}$: Count 
  of records in $T$ misclassified by $\mathcal{B}$ but not by $\mathcal{A}$.}
  &\multicolumn{1}{m{3cm}}{$n_{11}$: Count of records in $T$ misclassified by
  neither $\mathcal{A}$ nor $\mathcal{B}$.} \\
  \bottomrule
  \end{tabular}
  \end{table}

On the basis of $\mathcal{C}$, the conventional McNemar's test uses the 
following statistic \cite{RN1176}:
\begin{equation}
\mathcal{M}^{\textrm{HO}}=\frac{(|n_{01}-n_{10}|-1)^2}{n_{01}+n_{10}}
\sim \chi^2(1), \label{eqn:mcnemar_staitstic_ho}
\end{equation}
where $-1$ in the numerator is a continuity correction \cite{RN1205}. If 
$\mathcal{M}^{\textrm{HO}}>\chi^2_{\alpha}(1)$ where $\alpha$ is the 
significance level, then $H_0$ is rejected.

\subsection{Bayesian interpretation of $\mathcal{C}$}

For a record $z$, let $y$ be the true class label, and let $\hat{y}_\mathcal{A}$
and $\hat{y}_\mathcal{B}$ be the predicted class labels of classifiers 
$\mathcal{A}(S)$ and $\mathcal{B}(S)$, respectively. Then, a contingency table
$\mathcal{C}$ can be assumed to be a sample drawn from a multi-nomial 
distribution with parameters $\boldsymbol{\pi}=(\pi_{00},\pi_{01},\pi_{10},
\pi_{11})$ such that $\pi_{00}+\pi_{01}+\pi_{10}+\pi_{11}=1$. Formally, we 
denote $\mathcal{C}|\boldsymbol{\pi}\sim \mathbf{M}(n_2,\boldsymbol{\pi})$, or
\begin{equation}
P(\mathcal{C}|\boldsymbol{\pi}) = \frac{n_2!}{n_{00}!n_{01}!n_{10}!n_{11}!}
\pi_{00}^{n_{00}}\pi_{01}^{n_{01}}\pi_{10}^{n_{10}}\pi_{11}^{n_{11}}, 
\label{eqn:multi_nomial}
\end{equation}
where $\boldsymbol{\pi}$ is defined as follows:
\begin{eqnarray}
\pi_{00}&=&P(\hat{y}_\mathcal{A}\neq y, \hat{y}_{\mathcal{B}}\neq y),\nonumber\\ 
\pi_{01}&=&P(\hat{y}_\mathcal{A}\neq y, \hat{y}_{\mathcal{B}}=y), \nonumber \\
\pi_{10}&=&P(\hat{y}_\mathcal{A}=y, \hat{y}_{\mathcal{B}}\neq y), \nonumber \\
\pi_{11}&=&P(\hat{y}_\mathcal{A}=y, \hat{y}_{\mathcal{B}}=y). \nonumber
\end{eqnarray}

From $\boldsymbol{\pi}$, several meaningful random variables are derived, 
including a disagreement probability $e$ and three conditional error rates $r$, 
$q_a$, and $q_b$. Correspondingly, the estimators of these random variables are
induced from $\mathcal{C}$, namely, $E$, $R$, $Q_a$, and $Q_b$. Interpretation
of these random variables and estimators are presented in Table 
\ref{tbl:prob_notations}. Because $\mu_\mathcal{A}=\pi_{00}+\pi_{01}$ and 
$\mu_\mathcal{B}=\pi_{00}+\pi_{10}$, the problem in Eq. 
(\ref{eqn:core_problem}) can be rewritten as 
\begin{equation}
H_0:\ r = 0.5\ \ vs.\ \ H_1:\ r\neq 0.5.
\end{equation}

\begin{table*}[t]
\centering
\caption{Interpretation of several meaningful random variables defined on 
$\boldsymbol{\pi}$ and their estimators.}
\label{tbl:prob_notations}
\begin{tabularx}{\textwidth}{ccccX}
\toprule
Notation & Definition & Probabilistic form & Estimator & Interpretation \\
\midrule
$e$ & $\pi_{01}+\pi_{10}$ & $p(\hat{y}_\mathcal{A} \neq \hat{y}_\mathcal{B})$ &
  $E=\frac{n_{01}+n_{10}}{n_2}$ & Disagreement probability of the predictions  
  between algorithms $\mathcal{A}$ and $\mathcal{B}$ \\
$r$ & $\frac{\pi_{01}}{\pi_{01}+\pi_{10}}$ & $p(\hat{y}_\mathcal{A}\neq y|
\hat{y}_\mathcal{A}\neq \hat{y}_\mathcal{B})$ & $R=\frac{n_{01}}{n_{01}+n_{10}}$
  & Conditional error rate of algorithm $\mathcal{A}$ over the disagree 
  predictions \\
$q_a$ & $\frac{\pi_{01}}{\pi_{01}+\pi_{11}}$ & $p(\hat{y}_\mathcal{A}\neq y|
\hat{y}_\mathcal{B}= y)$ & $Q_a=\frac{n_{01}}{n_{01}+n_{11}}$ & Conditional 
error rate of algorithm $\mathcal{A}$ when algorithm $\mathcal{B}$ is correct\\
$q_b$ & $\frac{\pi_{00}}{\pi_{10}+\pi_{00}}$ & $p(\hat{y}_\mathcal{A}\neq y|
\hat{y}_\mathcal{B}\neq y)$ & $Q_b=\frac{n_{00}}{n_{10}+n_{00}}$ & Conditional 
error rate of algorithm $\mathcal{A}$ when algorithm $\mathcal{B}$ is 
incorrect\\
\bottomrule
\end{tabularx}
\end{table*}
 
In the estimators $E$, $R$, $Q_a$, and $Q_b$, the conditional distributions of 
$n_{01}+n_{10}$, $n_{00}$, and $n_{01}$ are as follows.
\begin{lemma}
Given that $\mathcal{C}|\boldsymbol{\pi}\sim \mathbf{M}(n_2,\boldsymbol{\pi})$,
we obtain
\begin{eqnarray}
n_{01}+n_{10} & \sim & \mathbf{B}(n_2, e),\nonumber \\
n_{01}|n_{01}+n_{10} & \sim & \mathbf{B}(n_{01}+n_{10}, r),\nonumber \\
n_{01}|n_{01}+n_{11} & \sim & \mathbf{B}(n_{01}+n_{11}, q_a),\nonumber \\
n_{00}|n_{10}+n_{00} & \sim & \mathbf{B}(n_{01}+n_{00}, q_b),\nonumber
\end{eqnarray}
where $\mathbf{B}(\cdot,\cdot)$ represents a binomial distribution.
\end{lemma}

From a Bayesian perspective, priors of $e$, $r$, $q_a$, and $q_b$ are assumed to 
be a conjugate Beta distribution $Be(\lambda, \lambda)$. Therefore, 
the posterior distributions of $e$, $r$ $q_a$, and $q_b$ conditioned on 
$\mathcal{C}$ are presented in Lemma \ref{lemma_ho_post_dist}.

\begin{lemma}
\label{lemma_ho_post_dist}
With a Beta prior $Be(\lambda, \lambda)$, posterior distributions of $e$, $r$, 
$q_a$, and $q_b$ are
\begin{eqnarray}
e|\mathcal{C}&\sim&Be(n_{01}+n_{10}+\lambda, n_{00}+n_{11}+\lambda),\nonumber \\
r|\mathcal{C} &\sim& Be(n_{01}+\lambda, n_{10}+\lambda), \nonumber\\
q_a|\mathcal{C} &\sim& Be(n_{01}+\lambda, n_{11}+\lambda), \nonumber\\
q_b|\mathcal{C} &\sim& Be(n_{00}+\lambda, n_{10}+\lambda).\nonumber
\end{eqnarray}
\end{lemma}

\begin{corollary}
Under the non-informative Beta prior with $\lambda=1$, the modes of the 
posteriors of $e$, $r$, $q_a$, and $q_b$ are
\begin{eqnarray}
mode[e|\mathcal{C}] & = & E,\nonumber \\
mode[r|\mathcal{C}] & = & R,\nonumber \\
mode[q_a|\mathcal{C}] & = & Q_a,\nonumber \\
mode[q_b|\mathcal{C}] & = & Q_b.\nonumber
\end{eqnarray}
\end{corollary}

It is noted that the Bayesian interpretation of a contingency table is not a
main contribution of this study because a similar Bayesian interpretation of 
a confusion matrix has been developed for analyzing precision, recall, and 
$\textrm{F}_1$ score \cite{RN776}\cite{wang2016credible}\cite{wang2019bayes}. 

\section{Na\"ive $K$-Fold CV McNemar's Test}
\label{sec:naive_kfcv_mcn}

In this section, we introduce an intuitive way to integrate the McNemar's test 
with a CV method. Considering the immensely popular $K$-fold CV, it produces $K$
contingency tables denoted as $\mathcal{C}_k=(n_{00,k},n_{01,k},n_{10,k},
n_{11,k})$ with $k=1,\ldots, K$. Furthermore, according to Eq. 
(\ref{eqn:mcnemar_staitstic_ho}), $K$ McNemar's test statistics can be defined, 
namely, $\mathcal{M}^{\textrm{HO}}_1,\ldots,\mathcal{M}^{\textrm{HO}}_K$.
Furthermore, we sum up the $K$ statistics and construct the following statistic.
\begin{equation}
\mathcal{M}^{\textrm{KCV}}=\sum_{k=1}^K\mathcal{M}^{\textrm{HO}}_k 
= \sum_{k=1}^K\frac{(|n_{01,k}-n_{10,k}|-1)^2}{n_{01,k}+n_{10,k}}.
\end{equation}
If $\mathcal{M}^{\textrm{HO}}_1,\ldots,\mathcal{M}^{\textrm{HO}}_K$ are 
independent, then $\mathcal{M}^{\textrm{KCV}}\sim \chi^2(K)$. Therefore, if 
$\mathcal{M}^\textrm{KCV}>\chi^2_\alpha(K)$, then $H_0$ is rejected.

The above statistic is na\"ive because $\mathcal{M}_1,\ldots,\mathcal{M}_K$ are
correlated due to a large overlap between training sets in $K$-fold CV.
Thus, $\mathcal{M}^{\textrm{KCV}}$ is not an exact $\chi^2(K)$. In this study, 
the test on $\mathcal{M}^{\textrm{KCV}}$ is named a na\"ive  $K$-fold CV 
McNemar's test, and it is not recommended.

\section{$5\times 2$ BCV McNemar's Test}
\label{sec:5x2bcvmcn}

\subsection{$5\times 2$ BCV }

Previous studies recommended a $5\times 2$ CV with random partitions (RCV) in a 
comparison of two algorithms \cite{RN391}\cite{RN722}\cite{RN720}. However, the 
random partitions of a $5\times 2$ RCV probably lead to excessively overlapping 
training sets which would degrade the performance of a algorithm comparison
method. In contrast, a $5\times 2$ BCV regularizes the numbers of overlapping 
records between any two training sets to be identical, and hence the 
variance of a $5\times 2$ BCV estimator of the error rate is smaller than 
that of a $5\times 2$ RCV estimator \cite{RN524}\cite{RN486}. Therefore, a 
$5\times 2$ BCV provides a promising data partitioning schema for developing 
a novel algorithm comparison method. 

Formally, a $5\times 2$ BCV performs five repetitions of two-fold CV 
partitioning with five regularized partitions. Let 
$\mathbb{P}=\{(S_j, T_j)\}_{j=1}^5$ denote a partition set of a $5\times 2$ BCV
with regularized conditions of $|S_j|=|T_j|=n/2$ and $|S_j\cap S_{j'}|\approx 
n/4$ for $j\neq j'$. Each partition $(S_j, T_j)$ corresponds to a two-fold CV in
which $S_j$ and $T_j$ are used as the training set in a round-robin manner.

For constructing $\mathbb{P}$,  $D_n$ is first divided into eight equal-sized
sub-blocks $D_1,\ldots,D_8$. The sub-blocks are combined to form $\mathbb{P}$ 
with the heuristic rules in Table \ref{tbl:partition_set_5x2} that is extracted 
from the first five columns of a two-level orthogonal array $L_8(2^7)$ in the 
domain of Design of Experiments (DoE) \cite{RN1170}.

\begin{table}[t]
\centering
\caption{Partition set $\mathbb{P}$ of a $5\times 2$ BCV.}
\label{tbl:partition_set_5x2}
\begin{tabular}{ccc}
\toprule
$j$ & Fold $S$ & Fold $T$ \\
\midrule
1 &$D_1,D_2,D_3,D_4$&$D_5,D_6,D_7,D_8$\\
2 &$D_1,D_3,D_5,D_7$&$D_2,D_4,D_6,D_8$\\
3 &$D_1,D_2,D_5,D_6$&$D_3,D_4,D_7,D_8$\\
4 &$D_1,D_4,D_5,D_8$&$D_2,D_3,D_6,D_7$\\
5 &$D_1,D_3,D_6,D_8$&$D_2,D_4,D_5,D_7$\\
\bottomrule
\end{tabular}
\end{table}

\subsection{Contingency tables on $5\times 2$ BCV}

When algorithms $\mathcal{A}$ and $\mathcal{B}$ are compared in a $5\times 2$
BCV, a collection of ten contingency tables is obtained, namely, 
$\boldsymbol{\mathcal{C}}_{5\times2}=\{\boldsymbol{\mathcal{C}}^{(j)}\}_{j=1}^5=
\{(\mathcal{C}_1^{(j)},\mathcal{C}_2^{(j)})\}_{j=1}^5$,
where $\mathcal{C}_k^{(j)}=\left(n_{00,k}^{(j)},n_{01,k}^{(j)}, n_{10,k}^{(j)}, 
n_{11,k}^{(j)}\right)$ with $k=1,2$. Table $\mathcal{C}_1^{(j)}$ uses $S_j$ and 
$T_j$ as the training and validation sets. Table $\mathcal{C}_2^{(j)}$ uses 
$T_j$ and $S_j$ to train and validate an algorithm.

Furthermore, an averaged contingency table $\bar{\mathcal{C}}_{5\times 2} =  
(\bar{n}_{00},\bar{n}_{01},\bar{n}_{10},\bar{n}_{11})$ is obtained based on 
 $\boldsymbol{\mathcal{C}}_{5\times 2}$, and it satisfies $\bar{n}_{00}+
 \bar{n}_{01}+\bar{n}_{10}+\bar{n}_{11}=n/2$ and
\begin{eqnarray}
\bar{n}_{ii'} = \frac{1}{10}\sum_{j=1}^5\sum_{k=1}^2n_{ii',k}^{(j)}, 
\ \ \ \ \forall i,i'=0,1.
\end{eqnarray}

Correspondingly, the estimators of $e$, $r$, $q_a$, and $q_b$ in 
$\bar{\mathcal{C}}_{5\times 2}$ are defined as follows.
\begin{eqnarray}
E_{5\times 2} & = &\frac{2(\bar{n}_{01} + \bar{n}_{10})}{n}, \nonumber \\
R_{5\times 2}&=&\frac{\bar{n}_{01}}{\bar{n}_{01}+\bar{n}_{10}}, \nonumber \\
Q_{a,5\times 2}& = & \frac{\bar{n}_{01}}{\bar{n}_{01}+\bar{n}_{11}},\nonumber \\
Q_{b,5\times 2}&=&\frac{\bar{n}_{00}}{\bar{n}_{10}+\bar{n}_{00}}.\nonumber
\end{eqnarray}

The denominator of $E_{5\times 2}$ is a constant, and thus 
\begin{equation}
E_{5\times 2} = \sum_{j=1}^5\sum_{k=1}^2\frac{(n_{01,k}^{(j)}+ n_{10,k}^{(j)})}
{5n}=\sum_{j=1}^5\sum_{k=1}^2\frac{E_{k}^{(j)}}{10}, \label{eqn:AGG_form}
\end{equation}
where $E_{k}^{(j)}$ is the estimator of $e$ on $\mathcal{C}_k^{(j)}$.

\begin{theorem}
The variance of $E_{5\times 2}$ is expressed as follows.  
\begin{equation}
\textrm{Var}[E_{5\times 2}|e] = \frac{1+\rho_1+8\rho_2}{5n}e(1-e),
\label{eqn:var_e5x2bcv}
\end{equation}
where $\rho_1$ and $\rho_2$ have the following definition.
\begin{itemize}
\item Let $\sigma^2=\textrm{Var}\left[E_{k}^{(j)}\right]=2e(1-e)/n$ be the 
variance of an HO estimator of $e$;
\item $\rho_1=\textrm{Cov}\left[E_{1}^{(j)},E_{2}^{(j)}\right]/\sigma^2$ is the 
inter-group correlation coefficient of two HO estimators in $E^{(j)}$ in a 
two-fold CV.
\item $\rho_2=\textrm{Cov}\left[E_{k}^{(j)},E_{k'}^{(j')}\right]/\sigma^2$ is 
the intra-group correlation coefficient of two HO estimators of $e$ in different
two-fold CVs where $j\neq j'$ and $k,k'=1,2$.
\end{itemize}
\end{theorem}

Eq. (\ref{eqn:var_e5x2bcv}) illustrates that $\rho_2$ plays a more important 
role than $\rho_1$. Moreover, because $\rho_1$ and $\rho_2$ have relations with  
specific algorithm types and data set distributions, it is hard to derive 
closed-form expressions of $\rho_1$ and $\rho_2$. Nevertheless, we are able to 
derive certain useful bounds of $\rho_1$ and $\rho_2$, and the bounds are 
indispensible in the statistical inference procedure in a comparison of two 
algorithms. 

\subsection{Properties of $\rho_1$ and $\rho_2$}

Given that the five partitions in $5\times 2$ BCV are performed on a single
data set,  all $E_{k}^{(j)}$s on $\boldsymbol{\mathcal{C}}_{5\times 2}$ are 
correlated. Thus, $\rho_1$ and $\rho_2$ usually deviate from zero.

Several theoretical properties of $\rho_1$ and $\rho_2$ are elaborated as 
follows.
\begin{lemma}
   Assume that $E_{k}^{(j)}$ merely depends on the validation set and $n$, then
   $\rho_1=0$ and $\rho_2=1/2$.
\end{lemma}

Nevertheless, the correlation between $E_{k}^{(j)}$s is also affected by the 
training sets in $5\times 2$ BCV because the training sets has a large overlap. 
Furthermore, under a mild assumption, the bound in Theorem \ref{thm_rho_bound}  
holds.
\begin{theorem}
\label{thm_rho_bound}
Assume that $E_{k}^{(j)}$ depends on the data set size $n$, the validation set, 
and numbers of overlapping records between the training sets in a $5\times 2$ 
BCV, the correlation coefficients $\rho_1$ and $\rho_2$ satisfy the following
bound.
\begin{equation}
\rho_2 < (1+\rho_1)/2.
\end{equation} 
Furthermore, if $\rho_1=0$, then $\rho_2<1/2$.
\end{theorem}

Many previous studies have investigated $\rho_1$ and $\rho_2$ over a broad 
family of loss functions, different algorithms, and various data sets 
\cite{RN524}\cite{RN857}\cite{RN486}. These studies confirmed that the bound in 
Theorem \ref{thm_rho_bound} holds, and they showed the bound is loose.
Furthermore, these studies slightly tighten the bound to the following form. 
\begin{equation}
\rho_1 \leq 1/2 \ \ \textrm{and} \ \ 0\leq \rho_2\leq 1/2. 
\label{eqn:rho_bound}
\end{equation}
They further revealed that the tight bounds hold with a high probability. Thus, 
they recommended the bounds in Eq. (\ref{eqn:rho_bound}) in a practical 
comparison scenario of two algorithms. Figure \ref{fig:dist_rho} also verifies 
that most values of $\rho_1$ and $\rho_2$ satisfy the bounds in Eq. 
(\ref{eqn:rho_bound}).

\subsection{Effective contingency table on a $5\times 2$ BCV}

In this section, we introduce a notion of ``effective contingency table" to
approximate the posterior distributions of variables $e$, $r$, $q_a$, and $q_b$ 
conditioned on the ten correlated contingency tables in 
$\boldsymbol{\mathcal{C}}_{5\times 2}$. The effective contingency table is a 
single virtual contingency table that is denoted as $\mathcal{C}_{e}=(n_{00,e},
n_{01,e},n_{10,e},n_{11,e})$. Moreover, let $n_e=n_{00,e}+n_{01,e}+n_{10,e}+
n_{11,e}$ denote the effective validation set size of $\mathcal{C}_e$.

The estimators of $e$, $r$, $q_a$, and $q_b$ on $\mathcal{C}_e$ are defined as 
$E_e$, $R_e$, $Q_{a,e}$, and $Q_{b,e}$ with the following forms.
\begin{eqnarray}
E_e & = & \frac{n_{01,e}+n_{10,e}}{n_e}, \nonumber\\ 
R_e&=&\frac{n_{01,e}}{n_{01,e}+n_{10,e}}, \nonumber \\
Q_{a,e} & = & \frac{n_{01,e}}{n_{01,e}+n_{11,e}}, \nonumber \\
Q_{b,e} & = & \frac{n_{00,e}}{n_{10,e}+n_{00,e}}.\nonumber
\end{eqnarray}
Under certain constraints with regard to the modes of $e$, $r$, $q_a$, and $q_b$
conditioned on $\mathcal{C}_e$ and the variance of the estimator $E_e$, a
closed-form expression of $\mathcal{C}_e$ is given as follows.
\begin{theorem}
\label{theorem_ce}
With the non-informative Beta prior and under the constraints of
\begin{eqnarray}
mode[e|\mathcal{C}_e] & = &  E_{5\times 2}, \nonumber \\  
mode[r|\mathcal{C}_e] & = & R_{5\times 2}, \nonumber \\
mode[q_a|\mathcal{C}_e] & = & Q_{a,5\times 2}, \nonumber \\
mode[q_b|\mathcal{C}_e] & = &  Q_{b,5\times 2}, \nonumber \\
Var[E_e|e] & = & Var[E_{5\times 2}|e], \nonumber
\end{eqnarray}
the closed-form expression of $\mathcal{C}_{e}$ is
\begin{equation}
\mathcal{C}_{e}=\frac{10}{1+\rho_1+8\rho_2}\bar{\mathcal{C}}_{5\times 2}, 
\label{eqn:effect_conttab_form}
\end{equation}
where $n_{00,e}$, $n_{01,e}$, $n_{10,e}$, and $n_{11,e}$ are
\begin{equation}
n_{ii',e} = \frac{10\bar{n}_{ii'}}{1+\rho_1+8\rho_2}, \ \ \ \nonumber
\forall i,i'=0,1.
\end{equation}
\end{theorem}

In terms of Eq. (\ref{eqn:effect_conttab_form}), we obtain the effective 
validation set size is
\begin{equation}
  n_e =\frac{5n}{1+\rho_1+8\rho_2}.
\end{equation}

\subsection{$5\times 2$ BCV McNemar's test}

Conditioned on $\mathcal{C}_e$ and $\rho_1$ and $\rho_2$, analogous to 
$\mathcal{M}^{\textrm{HO}}$ in Eq. (\ref{eqn:mcnemar_staitstic_ho}), a 
chi-squared statistic is defined as follows.
\begin{equation}
\mathcal{M}^{\textrm{BCV}}_{\rho_1,\rho_2}=\frac{(|n_{01,e}-n_{10,e}|-1)^2}
{n_{01,e}+n_{10,e}}\sim \chi^2(1), \label{eqn:mcnbcv}
\end{equation}

An equivalent form of $\mathcal{M}^{\textrm{BCV}}_{\rho_1,\rho_2}$ is obtained 
by substituting Eq. (\ref{eqn:effect_conttab_form}) into Eq. (\ref{eqn:mcnbcv}).
\begin{equation}
\mathcal{M}^{\textrm{BCV}}_t= \frac{t(|\bar{n}_{01}-\bar{n}_{10}|-1/t)^2}
{\bar{n}_{01}+\bar{n}_{10}}\sim \chi^2(1), \label{eqn:mcn5x2cv}
\end{equation}
where $t=10/(1+\rho_1+8\rho_2)$.

However, $\mathcal{M}_{t}^{\textrm{BCV}}$ contains an unknown parameter 
$t$, and thus it can not be used in a significance test. With an increasing 
$t$, the statistic $\mathcal{M}^{\textrm{BCV}}_t$ becomes large, and the 
McNemar's test induced from $\mathcal{M}^{\textrm{BCV}}_t$ tends to be a 
liberal test that easily produces false positive conclusions. In contrast,
according to the ``conservative principle" in the task of algorithm comparison 
\cite{RN729}, a smaller value of $t$ is preferred. Because $t$ is a monotonic 
decreasing function with regard to $\rho_1$ and $\rho_2$. Considering the 
bounds in Eq. (\ref{eqn:rho_bound}), we set $\rho_1=\rho_2=0.5$ and thus 
$t=20/11$. The following test statistic is obtained.
\begin{equation}
\label{eqn:mcnb5x2cvstat}
    \mathcal{M}^{\textrm{BCV}}=\frac{20(|\bar{n}_{01}-\bar{n}_{10}|-11/20)^2}
{11(\bar{n}_{01}+\bar{n}_{10})}\sim \chi^2(1), 
\end{equation} 

In this study, $\mathcal{M}^{\textrm{BCV}}$ is named a $5\times 2$ BCV 
McNemar's test statistic. Furthermore, the decision rule in a $5\times 2$ BCV
McNemar's test is: If $\mathcal{M}^{\textrm{BCV}}>\chi^2_\alpha(1)$, then $H_0$
is rejected. Moreover, $\alpha=0.05$ and $\chi^2_{0.05}(1)\approx 3.841$ are 
used.

\section{Experimental Results and Analysis}
\label{sec:expr}

To validate the superiority of the proposed $5\times 2$ BCV McNemar's test, we 
formulate two research questions (RQs).
\begin{description}
\item[RQ1:] How are the correlation coefficients $\rho_1$ and $\rho_2$ in Eq. 
(\ref{eqn:var_e5x2bcv}) distributed?
\item[RQ2:] Is the $5\times 2$ BCV McNemar's test better than the existing 
comparison methods of two classification algorithms?
\end{description}

For comparing two classification algorithms, the state-of-the-art methods 
mainly contain 4 families of 17 different significance tests. The existing tests
and their recommended settings are as follows.
\begin{enumerate}[(I)]
\item \textbf{$t$-test family} (9 tests). 
\begin{enumerate}[(1)]
   \item Repeated HO (RHO) paired $t$-test with 15 repeated HOs and 
$n_1=2n/3$ \cite{RN391}.
   \item $K$-fold CV paired $t$-test with $K=10$ \cite{RN391}.
   \item $5\times 2$ CV paired $t$-test \cite{RN391}.
   \item Corrected RHO $t$-test with $n_1=9n/10$ and 15 repeated HOs 
   \cite{RN729}.
   \item Pseudo bootstrap test \cite{RN729}.
   \item Corrected pseudo bootstrap test \cite{RN729}.
   \item Corrected $10\times 10$ CV $t$-test \cite{RN377}.
   \item Combined  $5\times 2$ CV $t$-test \cite{RN720}.
   \item Blocked $3\times 2$ CV $t$-test \cite{RN857}.
\end{enumerate}
\item \textbf{$F$-test family} (2 tests) 
\begin{enumerate}[(1)]
  \item Combined $5\times 2$ CV $F$-test \cite{RN722}.
  \item Calibrated $5\times 2$ BCV $F$-test \cite{RN486}. 
\end{enumerate}
\item \textbf{$Z$-test family} (3 tests) 
\begin{enumerate}[(1)]
  \item Proportional test with $n_1=2n/3$ \cite{RN391}.
  \item Conservative $Z$-test \cite{RN729}.
  \item $K$-fold CV-CI $Z$-test with $K=10$ and ``out'' type of variance 
   estimator \cite{byl2020}.
\end{enumerate}
\item \textbf{McNemar's test family} (3 tests) 
\begin{enumerate}[(1)]
  \item Conventional HO McNemar's test with $n_1=2n/3$ \cite{RN391}.
  \item $K$-fold CV Na\"{i}ve McNemar's test with $K=10$.
  \item $5\times 2$ BCV McNemar's test.
\end{enumerate}
\end{enumerate}

We exclude the conservative $Z$-test, pseudo bootstrap test, and corrected 
pseudo bootstrap test in all experiments because Nadeau and Bengio found these
three tests are less powerful and more expensive than the corrected RHO 
$t$-test \cite{RN729}. 

\subsection{Experiments for RQ1}

To show the distribution of $\rho_1$ and $\rho_2$, 29 real-world UCI data 
sets and seven popular classification algorithms are used. All the real-world
UCI data sets are listed in Table \ref{tbl:dataset}. For each data set, the
record that contains missing values is omitted. The counts of records,
predictors and classes in each data set are showed in Table \ref{tbl:dataset}. 
The seven classification algorithms used in the experiments are elaborated as 
follows.
\begin{enumerate}[(1)]
\item {\bf Majority classifier}. Test records are assigned with the class 
label with a maximum frequency in a training set. The assignment rule is
independent to the input predictors.
\item {\bf Mean classifier}. For each class, a mean vector of the predictors 
is calculated. A test record is assigned to the class whose mean vector has the 
smallest euclidean distance to the instance. It can be considered as a special 
case of linear discriminative analysis with a shared covariance matrix whose
diagonals are equal and off-diagonals are zero.
\item {\bf logistic regression.} The logistic regression classifier in ``RWeka"
package is used.
\item {\bf SVM.} The support vector machine with sequential minimal optimization 
is considered. The ``SMO" classifier with the default setting in ``RWeka" 
package is used.
\item {\bf RIPPER.}  The ``JRip" classifier with the default setting in ``RWeka" 
package is used.
\item {\bf C4.5.} The ``J48" classifier with the default setting in ``RWeka" 
package is used.
\item {\bf KNN.} The k-nearest neighborhood classifier in ``kknn" package is 
used. Parameter $k$ is set to 5, and a triangle kernel function is considered.
\end{enumerate}

\begin{table}[!t]
  \centering
  \caption{Data sets used in the experiments in the submitted paper.}
  \label{tbl:dataset}
  \begin{tabular}{ccccc}
  \toprule
  No. & Name & \#Record &\#Predictor & \#Class\\
  \midrule
  1&artificial&30654&7&10\\
  2&australian&690&14&2\\
  3&balance&625&4&3\\
  4&car&1728&6&4\\
  5&cmc&1473&9&3\\
  6&credit&690&15&2\\
  7&donors&748&4&2\\
  8&flare&1388&10&2\\
  9&german&1000&20&2\\
  10&krvskp&3196&36&2\\
  11&letter&20000&16&26\\
  12&magic&19020&10&2\\
  13&mammographic&830&5&2\\
  14&nursery&12960&8&5\\
  15&optdigits&5620&64&10\\
  16&page\_block&5473&10&5\\
  17&pendigits&10992&16&10\\
  18&pima&768&8&2\\
  19&ringnorm&7400&20&2\\
  20&satellite47&4435&36&6\\
  21&spambase&4601&57&2\\
  22&splice&3190&60&3\\
  23&tic\_tac\_toe&958&9&2\\
  24&titanic&2201&3&2\\
  25&transfusion&748&4&2\\
  26&twonorm&7400&20&2\\
  27&wave&5000&40&3\\
  28&wine\_quality&4898&11&7\\
  29&yeast&1484&7&10\\
  \bottomrule
  \end{tabular}
  \end{table}

On each data set, any two algorithms are compared, and thus $\binom{7}{2}=21$ 
different pairwise comparisons of algorithms are made. In total, 
$21\times29=609$ experiments are conducted. In each experiment, we randomly 
sample 300 records from a data set without replacement and then perform the 
block-regularized partitions on the records and compute the estimations 
$E_i^{(j)}$. The process is repeated in 1,000 times. The values of $\rho_1$ and 
$\rho_2$ are obtained over the 1,000 repetitions.   

\begin{figure}
  \centering
  \includegraphics[width=0.4\textwidth]{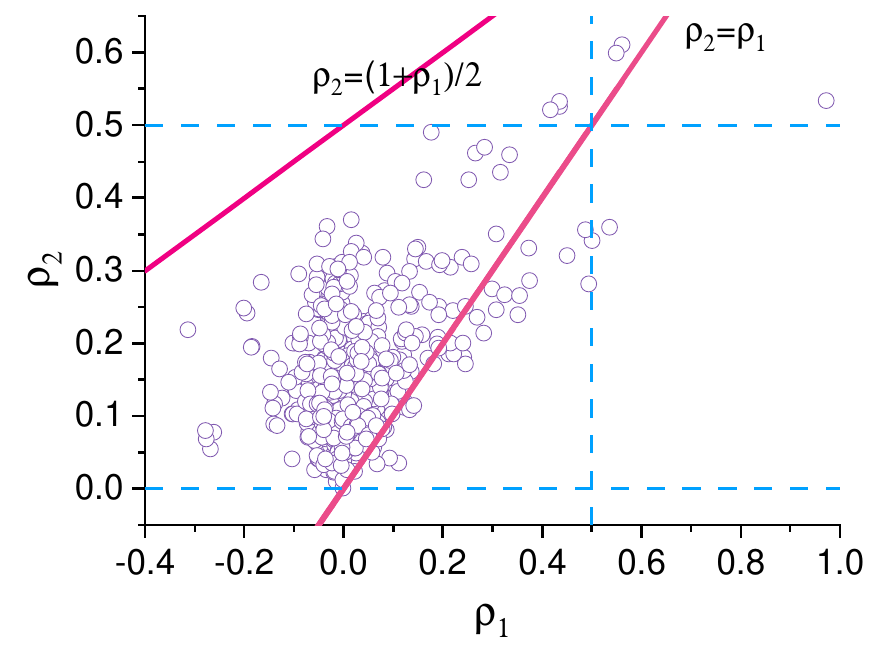}
  \caption{Scatter plot of $\rho_1$ and $\rho_2$.}
  \label{fig:dist_rho}
\end{figure}

Figure \ref{fig:dist_rho} shows the distribution of  the numeric values of 
$\rho_1$ and $\rho_2$ over the 609 experiments. Three observations are obtained 
as follows.  
\begin{enumerate}[(1)]
  \item Most values of $\rho_1$ and $\rho_2$ are distributed lower than 0.5. 
  Moreover, $\rho_1$ varies around 0, and $\rho_2$ is greater than 0. Therefore, 
  the bounds in Eq. (\ref{eqn:rho_bound}) hold in a high probability. The 
  observation is consistent to the previous studies.
  \item All values of $\rho_1$ and $\rho_2$ are located at the right lower area 
  of the red solid line $\rho_2=(1+\rho_1)/2$. The observation confirms the 
  bound in Theorem \ref{thm_rho_bound}. Furthermore, there is a wide margin 
  between the solid line and the scatter points, which indicates that the bound 
  in Theorem \ref{thm_rho_bound} is loose.
  \item Majority values of $\rho_1$ and $\rho_2$ satisfy $\rho_2 \geq \rho_1$. 
\end{enumerate}

\subsection{Experiments for RQ2}
\label{sec:exp_rq2}

Three synthetic data sets and two real-world data sets are used to answer the 
RQ2.

\textbf{Synthetic data set 1: Epsilon data set}. The epsilon data set is 
developed by Dietterich \cite{RN391} to investigate the type I error of a 
hypothesis test for comparing two classification algorithms. In the epsilon data 
set, on the basis of a data set size $n$, we directly generate the one-zero loss 
vectors of two algorithms. Specifically, let $i$ be a data record index where 
$i=1,\ldots,n$. Let $\mathbf{1}_{\mathcal{A}}(i)$ and 
$\mathbf{1}_{\mathcal{B}}(i)$ be the loss values of algorithms $\mathcal{A}$
and $\mathcal{B}$ on the $i$-th data record, respectively. Then, the loss values 
are generated according to the following rules.
\begin{itemize}
  \item When $1\leq i \leq n/2$, $\mathbf{1}_{\mathcal{A}}(i)
  \sim\mathbf{B}(1,\epsilon/2)$ and $\mathbf{1}_{\mathcal{B}}(i)
  \sim\mathbf{B}(1,3\epsilon/2)$.
  \item When $n/2+1\leq i \leq n$, $\mathbf{1}_{\mathcal{A}}(i)
  \sim\mathbf{B}(1,3\epsilon/2)$ and $\mathbf{1}_{\mathcal{B}}(i)
  \sim\mathbf{B}(1,\epsilon/2)$. 
\end{itemize}

According to the rules, we can obtain that on the first half of the data set, 
the error rates of algorithms $\mathcal{A}$ and $\mathcal{B}$ are 
$\mu_{\mathcal{A}}=\epsilon/2$ and 
$\mu_{\mathcal{A}}=3\epsilon/2$; on the remaining half of the data set,
$\mu_{\mathcal{A}}=3\epsilon/2$ and 
$\mu_{\mathcal{A}}=\epsilon/2$; and on the entire data set, we obtain 
$\mu_{\mathcal{A}}=\mu_{\mathcal{B}}=\epsilon$, and thus the null hypothesis
$H_0$ in Eq. (\ref{eqn:core_problem}) is true. Hence, the epsilon data set can 
merely be used to obtain the type I error of a test. In this data set, the
settings of $n=300$ and $\epsilon=0.1$ are used.  

\begin{figure}
  \centering
  \includegraphics[width=0.45\textwidth]{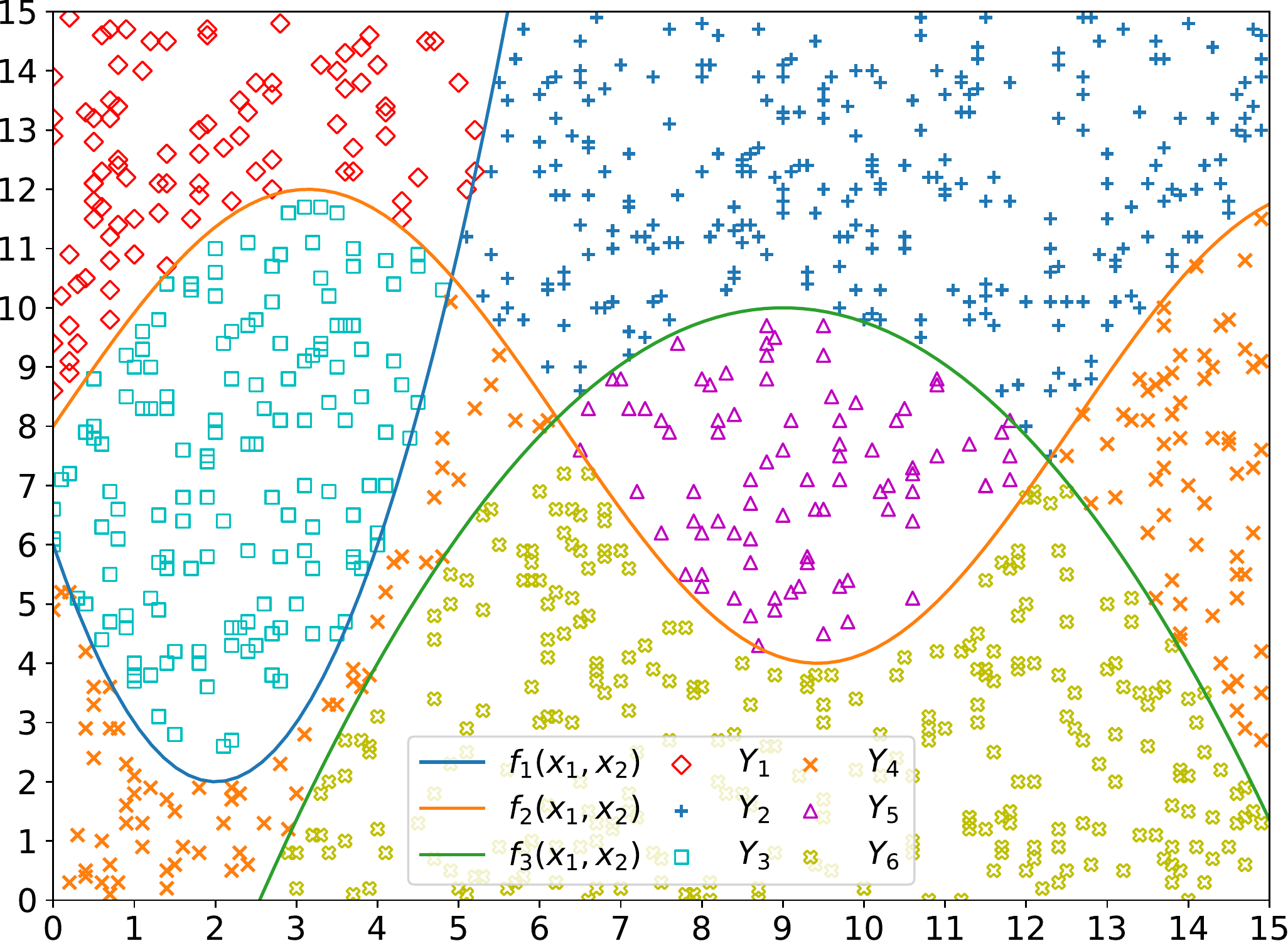}
  \caption{Demonstration of an EXP6 data set with $n=300$.}
  \label{fig:exp6plot}
\end{figure}

\textbf{Synthetic data set 2: EXP6 data set}. The EXP6 data set is a six-class 
learning problem with two continuous predictors. It was developed in 
\cite{kong1995error} and used in the algorithm comparison task in \cite{RN391}. 
Specifically, in an EXP6 data set $D_n=\{(y_i, \boldsymbol{x}_i)\}_{i=1}^n$, 
let $\boldsymbol{x}=(x_1, x_2)$ be two predictors and $y$ be a true class label 
satisfying $y\in \{Y_j\}_{j=1}^6$. Predictors $x_1$ and $x_2$ in an EXP6 data 
set are uniformly distributed over an uniform grid space of resolution 0.1 over 
the region $\mathbf{X}=[0, 15]\times [0, 15]$. Class label $y_{\mathbf{x}}$ is 
generated as follows.
\begin{equation}
y_{\mathbf{x}} = \left\{
\begin{aligned}
Y_1 &\ \ \ \   x_2-f_1(x_1) \geq 0\ \land\ x_2-f_2(x_1)\geq 0 \\
Y_2 &\ \ \ \  x_2-f_1(x_1) < 0 \ \land \ x_2-f_2(x_1)\geq 0 \\
&\ \ \ \  \land\ x_2-f_3(x_1) \geq 0\\
Y_3 &\ \ \ \  x_2-f_1(x_1) \geq 0\ \land\ x_2-f_2(x_1)<0 \\
Y_4 &\ \ \ \  x_2-f_1(x_1) < 0 \ \land\ x_2-f_2(x_1)< 0 \\
&\ \ \ \  \land\ x_2-f_3(x_1) \geq 0\\
Y_5 &\ \ \ \   x_2-f_2(x_1) \geq 0\ \land\ x_2-f_3(x_1)< 0 \\
Y_6 &\ \ \ \   x_2-f_2(x_1) < 0\ \land\ x_2-f_3(x_1)< 0 \\
\end{aligned}
\right.
\end{equation}
where $f_1$, $f_2$ and $f_3$ are the following decision boundary functions.
\begin{eqnarray}
f_1(x) & = & x^2 -4x+6,\nonumber \\
f_2(x) & = & 4sin(x/2)+8,\nonumber \\
f_3(x) & = & -\frac{1}{25}(x^2-108x+236).
\end{eqnarray}
A scatter plot of an EXP6 data set with $n=300$ is showed in Figure 
\ref{fig:exp6plot}. Furthermore, two algorithms are compared on an EXP6 data 
set.
\begin{LaTeXdescription} 
\item[($\mathcal{A}$) \textbf{C4.5}.] It is a popular tree-based classification 
algorithm. J48 in package RWeka implements a C4.5 algorithm and is used with no
pruning procedure.
\item[($\mathcal{B}$) \textbf{FNN}.] First nearest neighbor uses a weighted 
Euclidean distance to tune its performance to the level of the C4.5.
Specifically, the distance between points $\boldsymbol{x}_{i}=(x_{i1}, x_{i2})$ 
and $\boldsymbol{x}_{j}=(x_{j1}, x_{j2})$ is 
\begin{equation}
d(\boldsymbol{x}_i, \boldsymbol{x}_j) = \omega (x_{i1}-x_{j1})^2 + 
\omega^{-1}(x_{i2}-x_{j2})^2,
\end{equation}
where $\omega\in (0.0, 1.0]$ is a tunable weight. When $\omega=1.0$, 
$d(\cdot,\cdot)$ degrades to the typical Euclidean distance.  
\end{LaTeXdescription}

\begin{figure*}[!ht]
  \centering
  \includegraphics[width=0.48\textwidth]{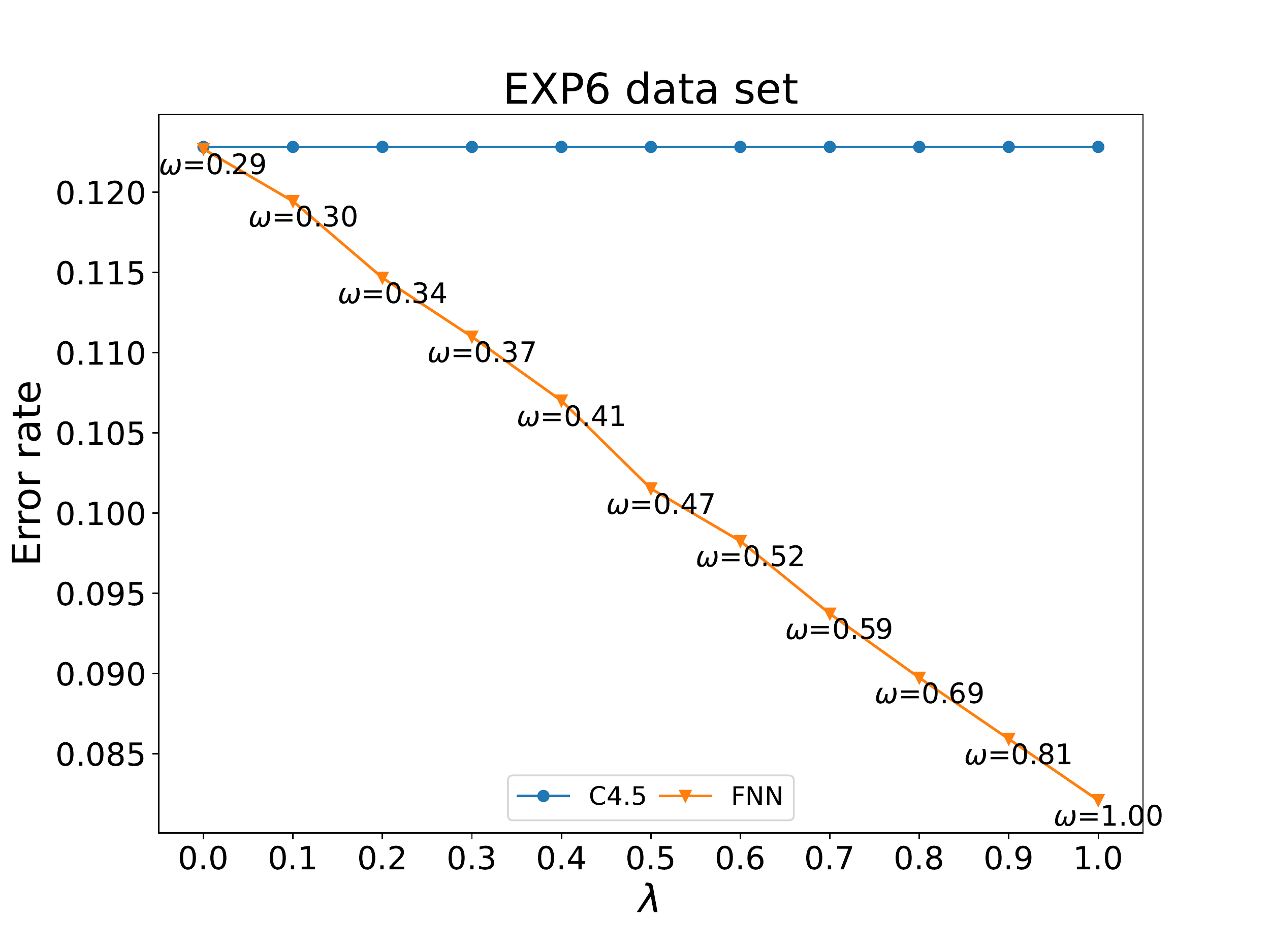}
  \includegraphics[width=0.48\textwidth]{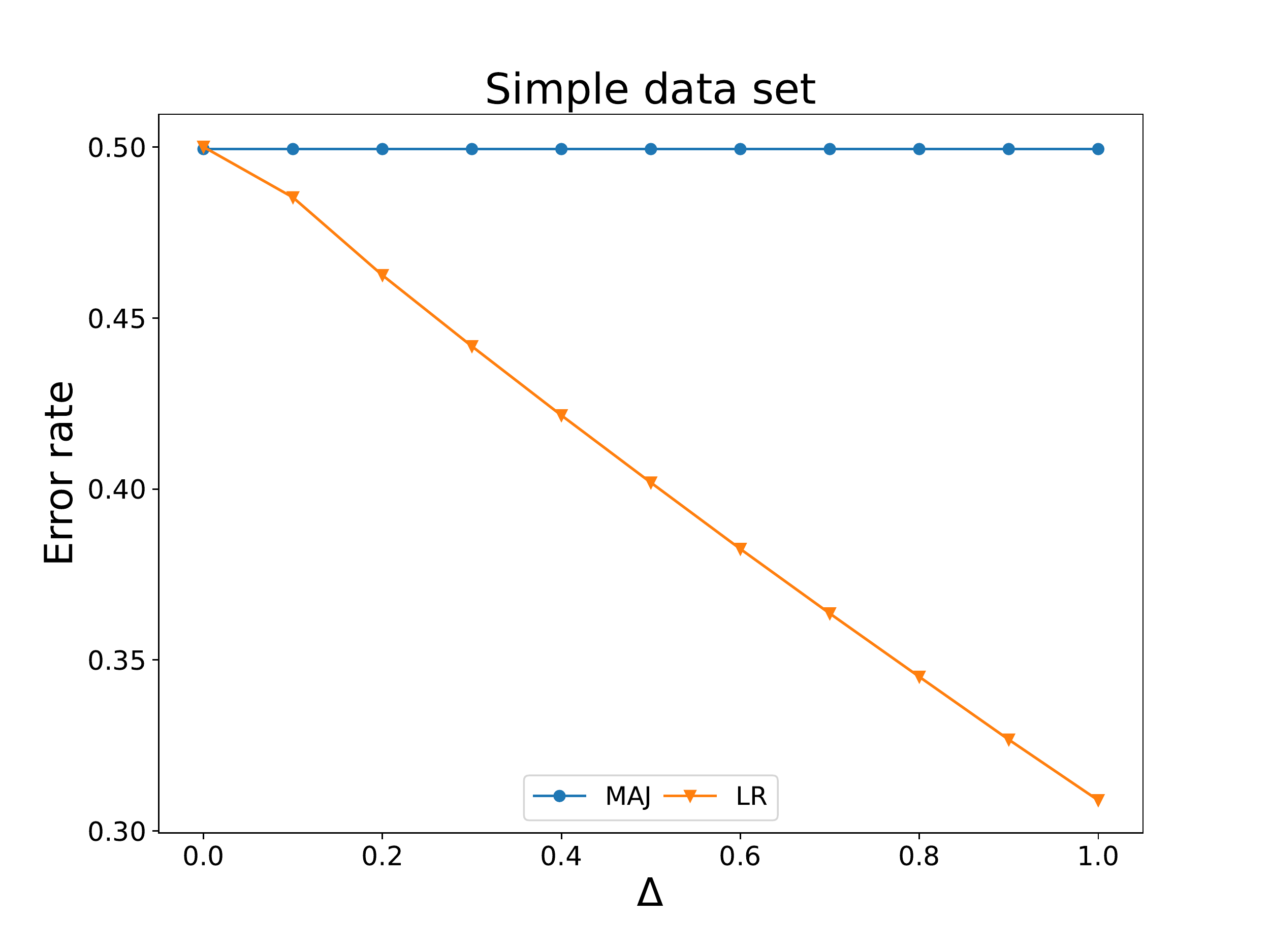}
  \includegraphics[width=0.48\textwidth]{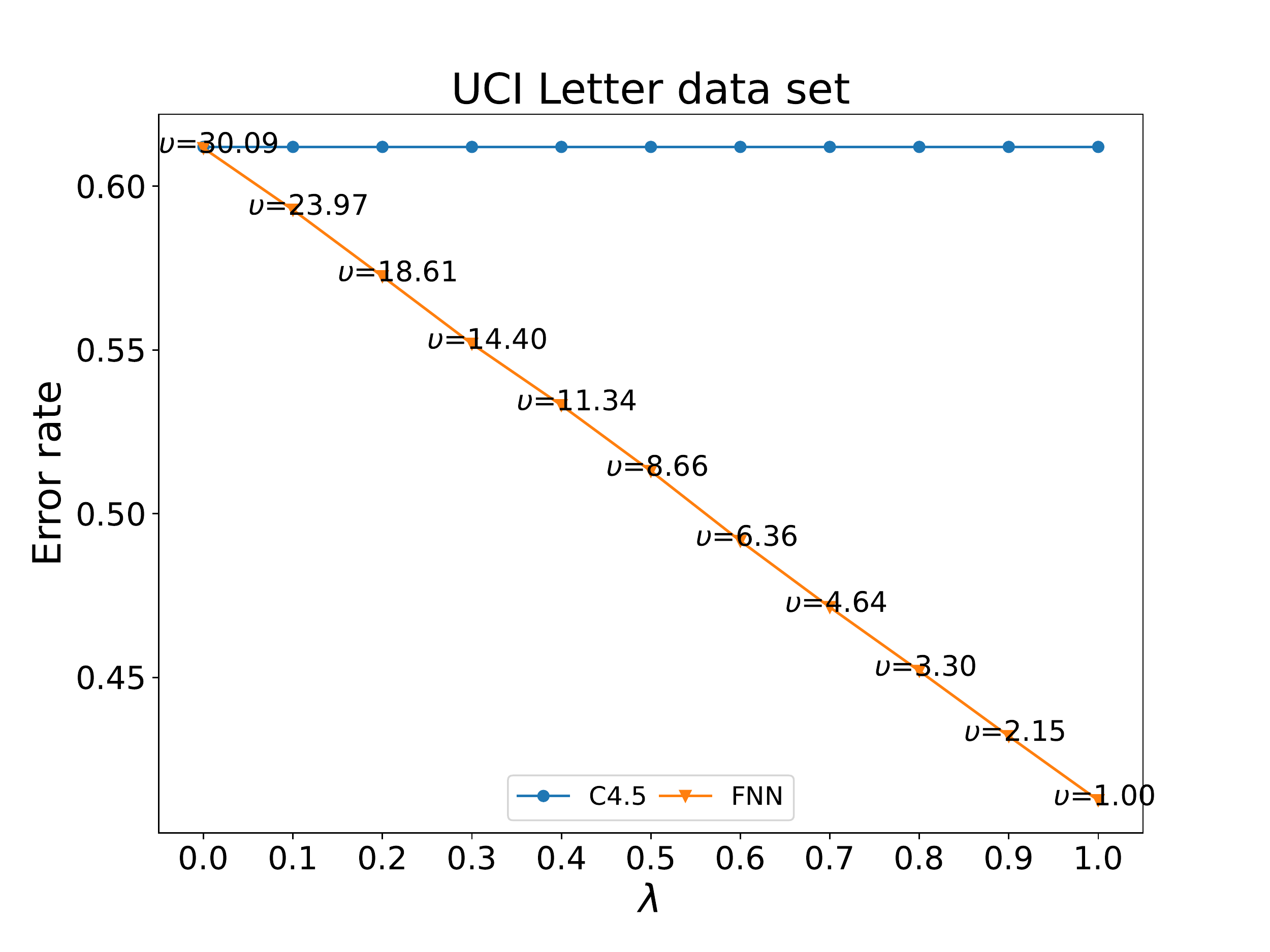}
  \includegraphics[width=0.48\textwidth]{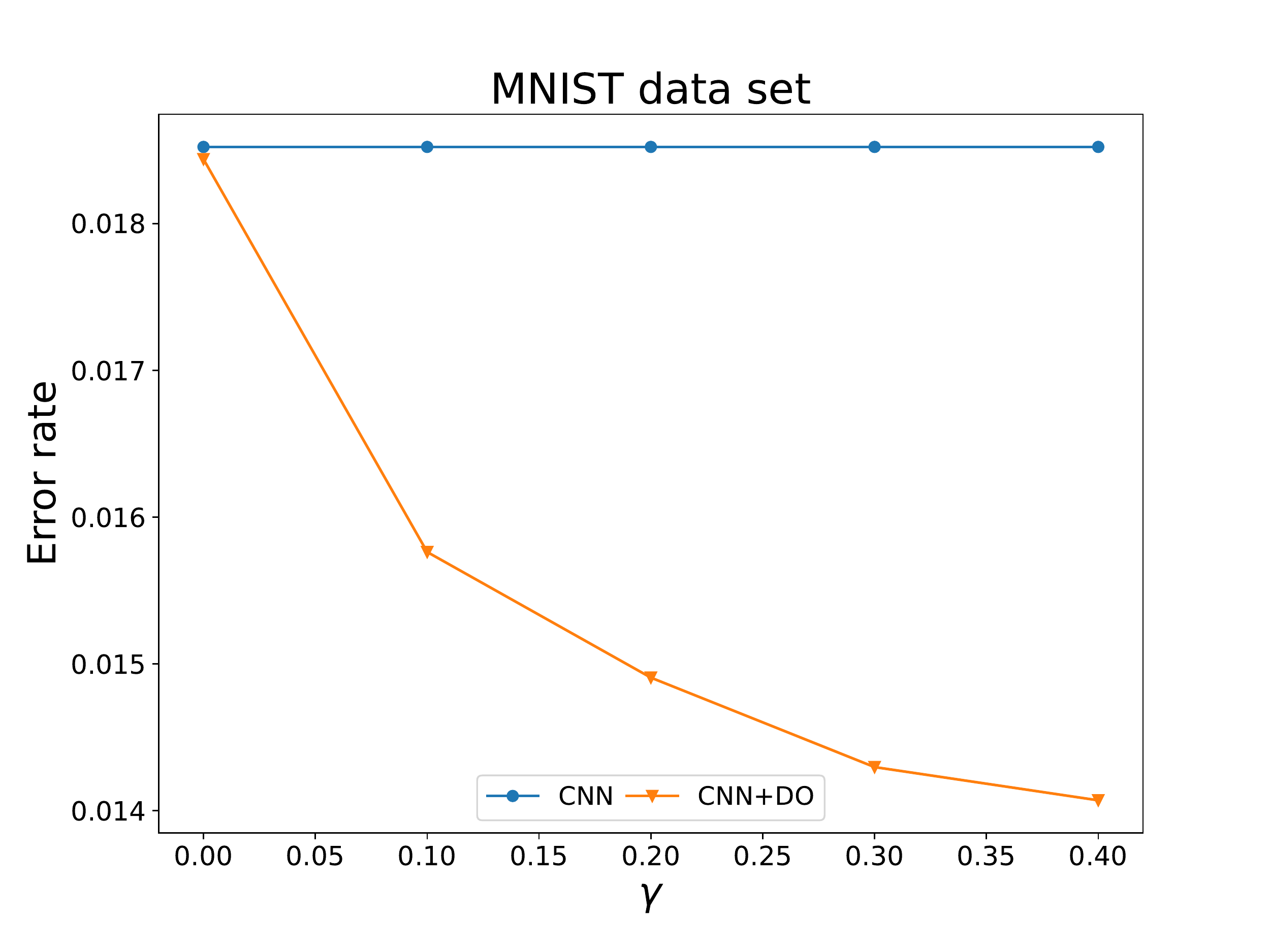}
  \caption{Simulations of true error rates of algorithms on the synthetic and
real-world data sets}
  \label{fig:true_error_rate}
\end{figure*}

\begin{table*}[t]
\centering
\caption{Type I errors of different algorithm comparison methods over synthetic
and real-world data sets.}
\label{tbl:typeierrors}
\begin{tabular}{ccccccc}
\toprule
\multirow{2}{*}{Family} & \multirow{2}{*}{Test} & 
\multicolumn{3}{c}{Synthetic} & \multicolumn{2}{c}{Real-world}\\
\cmidrule(lr){3-5}\cmidrule(lr){6-7}
&& Epsilon & EXP6 & Simple & UCI letter & MNIST \\
\midrule
\multirow{7}{*}{$t$-test family}&RHO paired $t$-test & 0.478 & 0.272 & 0.312 & 
0.385 & 0.250  \\
&$K$-fold CV paired $t$-test & \textbf{0.043} & 0.076 & 0.109 & 0.142&0.090  \\
&$5\times 2$ CV paired $t$-test & \textbf{0.034} & \textbf{0.050} & 0.084 & 0.061
& 0.080 \\
&Corrected RHO $t$-test & 0.053 & \textbf{0.040} & \textbf{0.047} & 0.075 &
\textbf{0.000}  \\
&Corrected $10\times 10$ CV $t$-test & \textbf{0.035} & \textbf{0.040} & 0.063 & 
0.082 & 0.080  \\
&Combined $5\times 2$ CV $t$-test & 0.291 & 0.156 & 0.214 & 0.236 & 0.150 \\
&Blocked $3\times 2$ CV $t$-test & 0.087 & \textbf{0.005} & \textbf{0.015} & 
\textbf{0.013} &\textbf{0.020}  \\
\cmidrule(lr){2-7}
\multirow{2}{*}{$F$-test family}&Combined $5\times 2$ CV $F$-test & 
\textbf{0.028} & \textbf{0.028} & 0.060 & 
0.057 & \textbf{0.010}  \\
&Calibrated $5\times 2$ BCV $F$-test & \textbf{0.035} & \textbf{0.025} & 0.055 & 
\textbf{0.042} & \textbf{0.040} \\
\cmidrule(lr){2-7}
\multirow{2}{*}{$Z$-test family}&Proportional test & 0.056 & \textbf{0.016} & 
\textbf{0.014} & 0.054 & \textbf{0.020} \\
&$K$-fold CV-CI $Z$-test & \textbf{0.046} & 0.062 & 0.106 & 0.080 & 0.090  \\
\cmidrule(lr){2-7}
\multirow{3}{*}{McNemar's test family}&Conventional HO McNemar's test & 
\textbf{0.031} & \textbf{0.037} & \textbf{0.029} & 0.062 & 0.060  \\
&Na\"{i}ve $K$-fold CV McNemar's test & \textbf{0.000} & \textbf{0.006} & 
\textbf{0.020} & \textbf{0.039}  & \textbf{0.020}  \\
&$5\times 2$ BCV McNemar's test & \textbf{0.025} & \textbf{0.006} & 
\textbf{0.005} & \textbf{0.015} & \textbf{0.010}  \\
\bottomrule
\end{tabular}
\end{table*}

\begin{figure*}[!ht]
  \centering
  \includegraphics[width=0.47\textwidth]{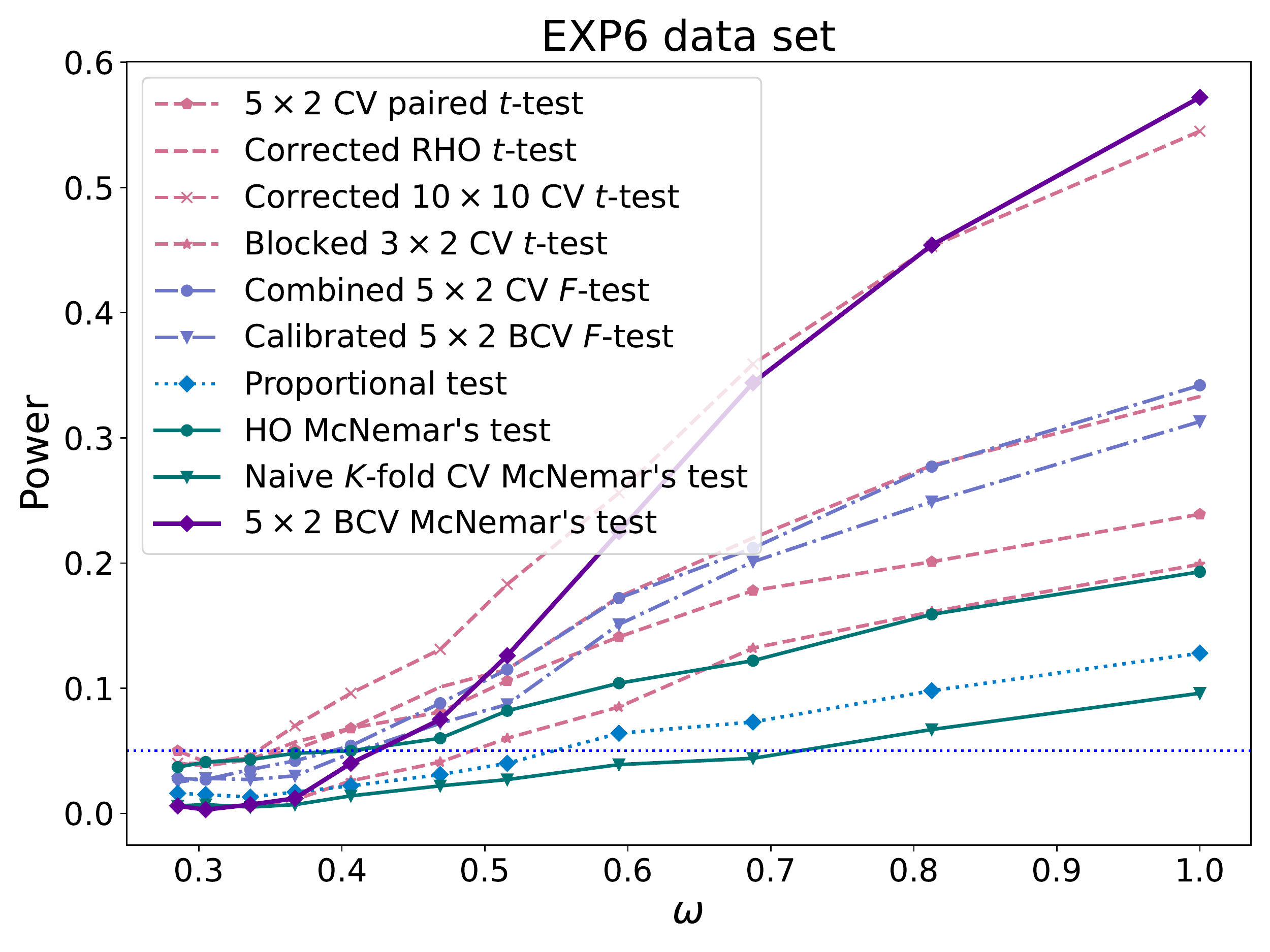}
  \includegraphics[width=0.47\textwidth]{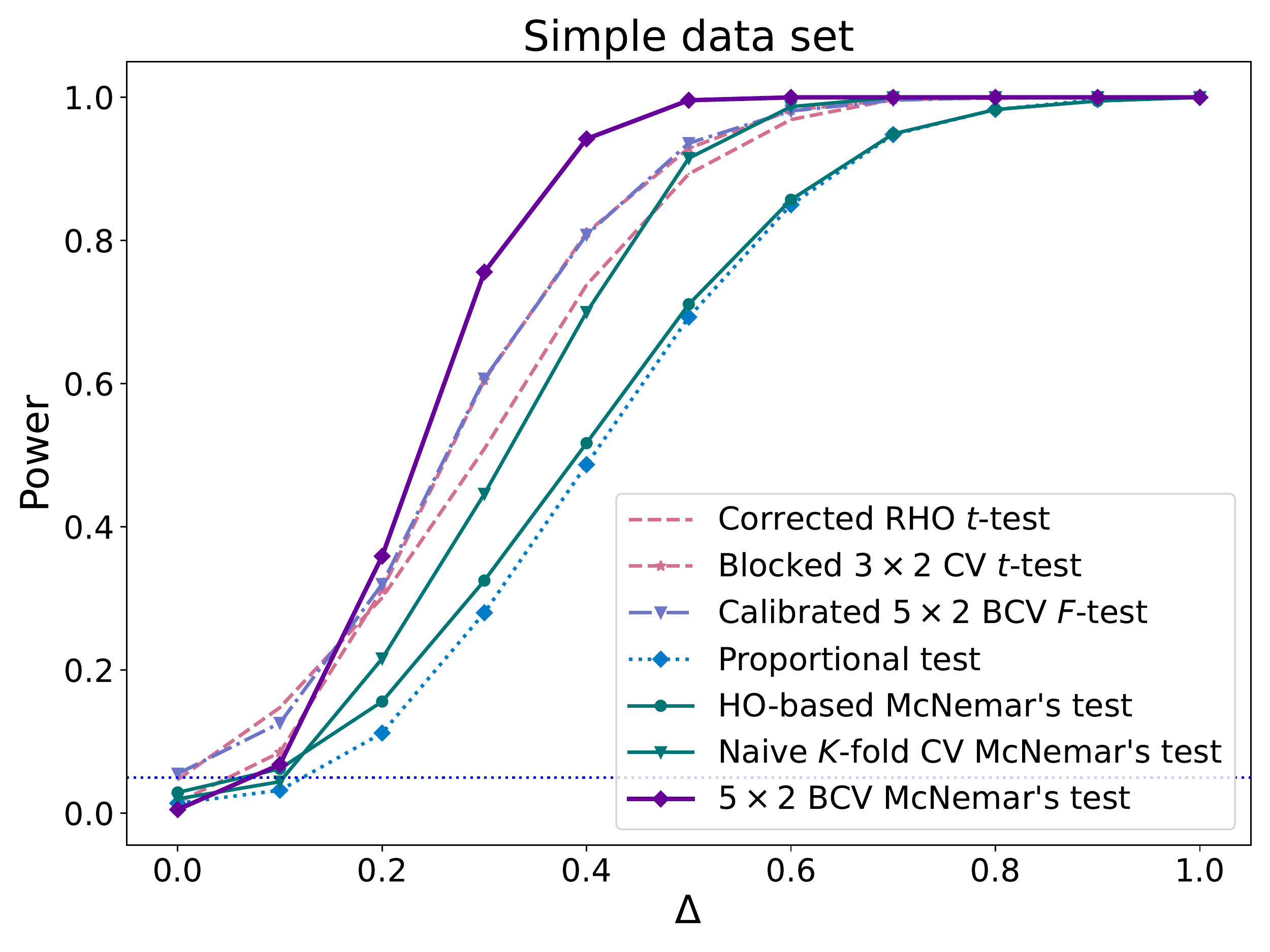}
  \includegraphics[width=0.47\textwidth]{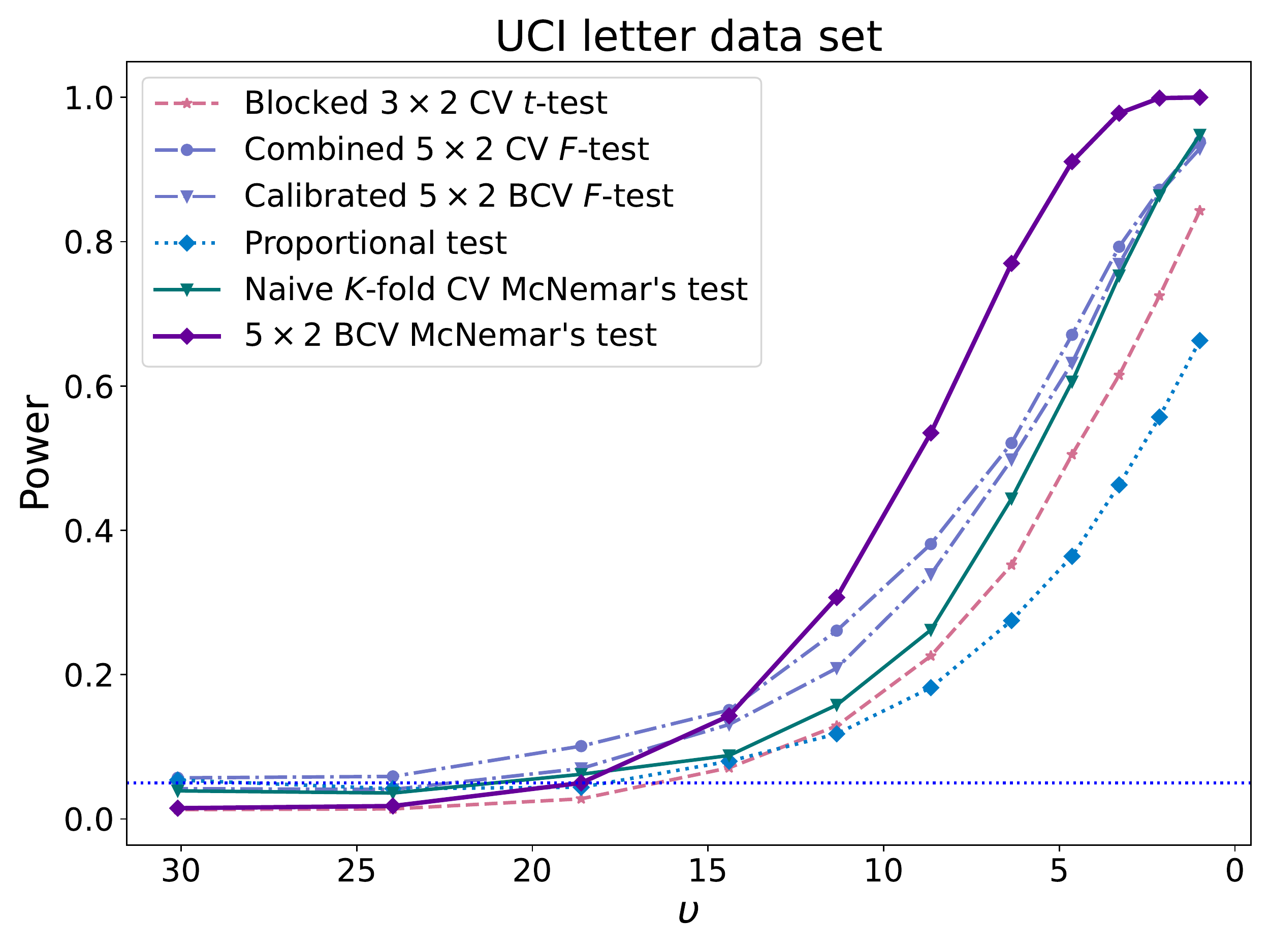}
  \includegraphics[width=0.47\textwidth]{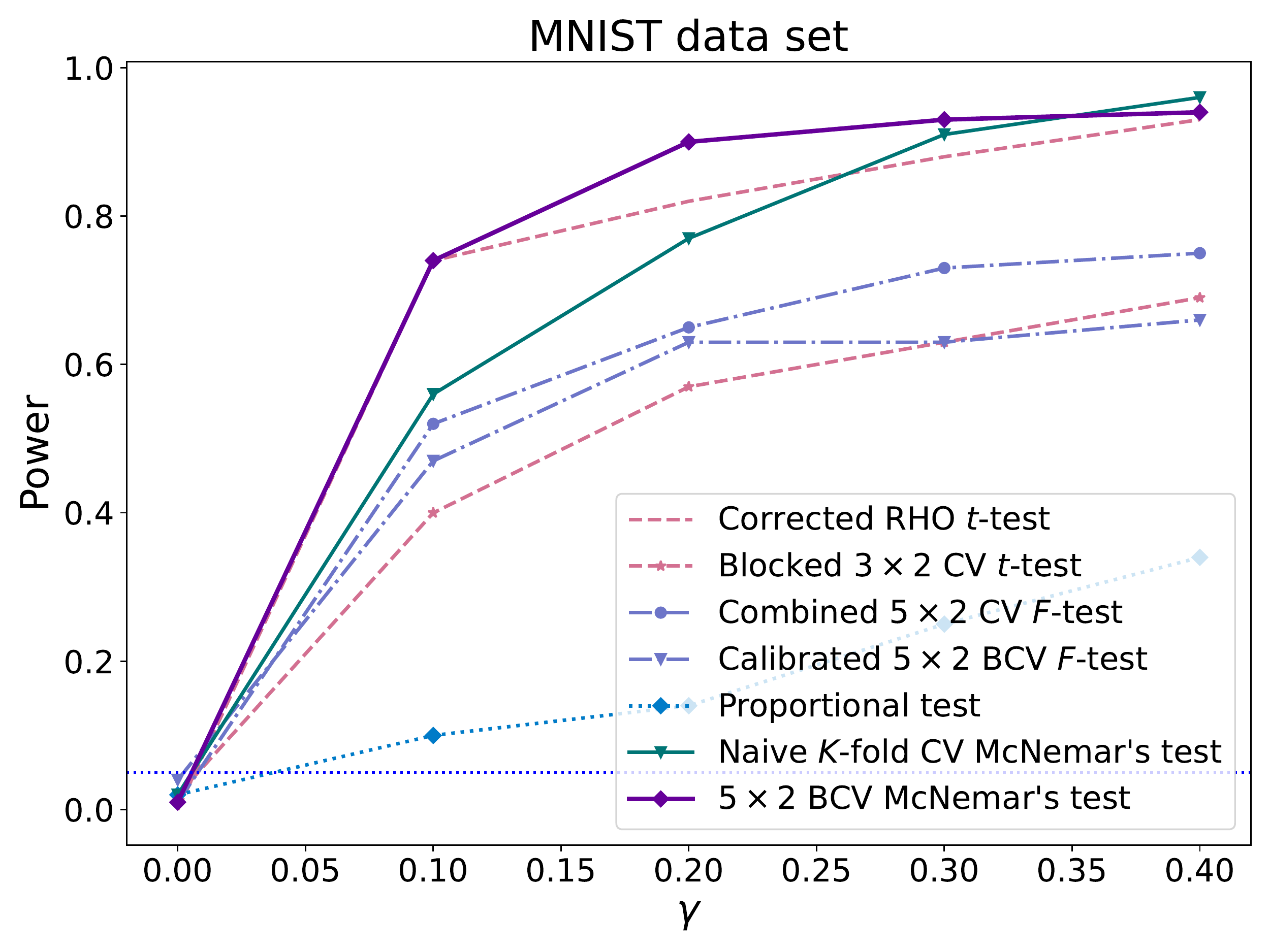}
  \caption{Power curves of different tests on the synthetic and real-world data
 sets.}
  \label{fig:power_plots}
\end{figure*}

True values of the error rates of C4.5 and FNN, namely, $\mu_{C4.5}$ and 
$\mu_{FNN,\omega}$, are obtained through taking averages over all estimates of
the error rates of the classifiers that are trained over 1,000 iid EXP6 data 
sets with $n=300$ and evaluated with a large EXP6 data set with 
$(15/0.1 +1)^2=22,801$ records. Because the error rate of FNN depends on the 
hyper-parameter $\omega$, we introduce multiple target error rates for FNN 
algorithm. Specifically, a target error rate $\mu_{target}$ is expressed as a 
linear interpolation value between $\mu_{C4.5}$ and $\mu_{FNN,1.0}$, i.e., 
$\mu_{target}=\mu_{C4.5}-\lambda(\mu_{C4.5}-\mu_{FNN,1.0})$. We adjust $\omega$ 
so that the true error rate of an FNN achieves $\mu_{target}$. The discrete 
values of $\lambda$ are derived from the range of [0.0, 1.0] with a step of 0.1. 
In particular, when $\lambda=0.0$, the target error rate of an FNN degrades to 
$\mu_{C4.5}$, which makes the null hypothesis $H_0$ in Eq. 
(\ref{eqn:core_problem}) being true.   

Figure \ref{fig:true_error_rate} shows the true error rate of C4.5 and FNN with 
various target error rates. Several specific observations are obtained from 
Figure \ref{fig:true_error_rate}. The true error rate of C4.5 is 12.27\%. The
true error rate of FNN decreases with an increasing $\omega$. When $\omega=1.0$, 
the error rate of FNN is 8.21\%. When $\omega=0.29$, FNN achieves an identical 
error rate with C4.5. Therefore, type I error of a significance test method is 
obtained with the setting of $\omega=0.29$. Power curve of a test is depicted 
by increasing $\omega$ from 0.29 to 1.0.

\textbf{Synthetic data set 3: Simple data set}. Let 
$D_n=\{(x_i,y_i)\}_{i=1}^{n}$ be a simple binary classification data set 
where $y_i\in \{Y_0, Y_1\}$ is the true class label and $x_i$ is the predictor. 
Class label index $i\sim \mathbf{B}(1, 0.5)$ is independently drawn. 
Then, $x_i|y_i=Y_1\sim N(0, 1)$ and $x_i|y_i=Y_2\sim N(\Delta, 1)$ where 
$\Delta\in[0.0, 1.0]$ is a tunable parameter. Moreover, on a simple data set,
two algorithms are compared. 
\begin{LaTeXdescription}
\item[($\mathcal{A}$) \textbf{LR}.] Logistic regression is implemented by the
``glm" package in R software with a binomial link function. 
\item[($\mathcal{B}$) \textbf{MAJ}.] Majority classifier uses the most frequent 
class label in the training set as a prediction that is independent to the input
predictors.
\end{LaTeXdescription}

True error rates of LR and MAJ on a simple data set are illustrated in 
Figure \ref{fig:true_error_rate}. Regardless of $\Delta$, the error rate of MAJ 
remains unchanged because it merely depends on the prior 
distribution of class labels. Moreover, when $\Delta=0.0$, the distributions of
the predictor $x$ in two classes are completely overlapping, and thus LR 
degrades to a random guess that possesses an identical error rate with MAJ. When
$\Delta$ increases, the error rate of LR monotonically decreases. Therefore, 
$\Delta=0.0$ is used to obtain the type I error of a test, and a power curve is
depicted by increasing $\Delta$ from 0.0 to 1.0. In a simple data set, $n=1,000$ 
is used.

\textbf{Real-world data set 1: UCI letter data set}. It is a popular data set in
the task of algorithm comparison \cite{RN391,RN729}. It contains 20,000 data 
points, 26 classes, and 16 predictors. The following two classification 
algorithms are compared. 
\begin{LaTeXdescription}
\item[($\mathcal{A}$) \textbf{TREE}.] Classification tree uses the package 
``tree" with default settings in R software.
\item[($\mathcal{B}$) \textbf{FNN}.] First nearest neighbor uses a distorted 
distance $d(\cdot,\cdot)$ as follows \cite{RN729}.
\begin{equation}
d(\boldsymbol{x}_i, \boldsymbol{x}_j)=\sum_{k=1}^3\upsilon^{2-k}
\sum_{l\in C_k}(x_{il}-x_{jl})^2,
\end{equation}
where $\boldsymbol{x}_i$ and $\boldsymbol{x}_j$ are two predictor vectors, 
$C_1=\{1,3,9,16\}$, $C_2=\{2,4,6,7,8,10,12,14,15\}$, and $C_3=\{5,11,13\}$. 
$\upsilon\in [1, 50]$ is a tunable parameter to adjust the error rate of an FNN. 
\end{LaTeXdescription}

Furthermore, 1,000 data sets of $n=300$ are iid drawn from the letter data set 
with replacement. True values of the error rates of TREE and FNN are obtained 
by training the algorithms on the 1,000 data sets and validating on the entire 
letter data set. Similar to the EXP6 data set, the simulation of the true error
rates is based on several targeted error rates characterized with an 
interpolation parameter $\lambda$.  The true error rates are showed in Figure 
\ref{fig:true_error_rate} which illustrates that the error rate of FNN 
monotonically increases with an increasing $\upsilon$. When $\upsilon=30.09$, 
FNN owns an identical error rate with TREE. Thus, the type I error of a test is 
obtained when $\upsilon=30.09$, and the power curve of a test is depicted by 
decreasingly changing $\upsilon$ from $30.09$ to $1.0$.

\textbf{Real-world data set 2: MNIST data set}. It is a popular ten-class 
classification data set in deep learning studies. The gold split of the MNIST
data set in the keras library consists of a training set with 60,000 images and
a test set with 10,000 images. We merely use the 60,000 images in the training 
set as a data population. Two deep learning models are compared.
\begin{LaTeXdescription}
\item[($\mathcal{A}$) CNN.] The architecture of the used CNN is expressed as
conv$\to$max\_pooling$\to$conv$\to$max\_pooling$\to$flatten
$\to$dense$\to$softmax. We use ReLU as an activation function and cross-entropy 
as an objective function. The batch size and epoch count are set to 128 and 250.
Numbers of filters in the two ``conv" layers are 32 and 64, respectively. 
A same kernel size of $(5, 5)$ is used in the two ``conv'' layers. Moreover, a 
pooling size of $(2, 2)$ is used. Size of the hidden dense layer is 500, and 
size of the softmax layer is 10.
\item[($\mathcal{B}$) CNN+DO.] We apply dropout regularization to the weights
between the flatten layer and the dense layer of the above CNN. The dropout rate  
$\gamma \in [0.0, 1.0)$ is used to tune the error rate of CNN+DO.
\end{LaTeXdescription}

Considering that it is computationally expensive to train a CNN model, we only 
generate 100 data sets, each of size $n=10,000$, to simulate 100 
separate trails. We use the entire population as a test set to compute the true
error rates by averaging the 100 estimators of the models trained on the 100 
data set. We changes the drop out rate from 0.0 to 0.4 with a step of 0.1. The 
true values of the error rates of CNN and CNN+DO are plotted in the last plot of 
Figure \ref{fig:true_error_rate}. The true error rate of CNN is 1.85\% 
regardless of dropout rate $\gamma$. The true error rate of CNN+DO is decreasing 
with an increasing drop out rate $\gamma$. When $\gamma=0.0$, the error rate of 
CNN+DO is equivalent to that of CNN. When $\gamma$ increases, CNN+DO has a lower 
error rate than CNN. Therefore, for a significance test, we use $\gamma=0.0$ 
obtain its type I error, and we obtain its power curve by changing $\gamma$ 
from 0.0 to 0.4.

The type I errors of all significance tests over the simulated and real-world
data sets are presented in Table \ref{tbl:typeierrors}. The type I errors below
$\alpha=0.05$ is indicated in bold font. Two observations are obtained. 
\begin{enumerate}[(1)]
\item All type I errors of the $5\times 2$ BCV McNemar's test and na\"{i}ve 
$K$-fold CV McNemar's test are lower than 0.05. It indicates that the two tests 
can effectively reduce false positive conclusions for algorithm comparison. 
Majority of type I errors of blocked $3\times 2$ CV $t$-test and calibrated 
$5\times 2$ BCV $F$-test are reasonable. In contrast, although $K$-fold CV 
paired $t$-test and $K$-fold CV-CI $Z$-test own reasonable type I errors on the 
epsilon data set, all type I errors of the two tests on other data sets exceed
0.05. Therefore, the two tests are unpromising for comparing two classification 
algorithms. Moreover, all type I errors of the RHO paired $t$-test, combined 
$5\times 2$ CV $t$-test are larger than 0.05, and thus the two tests which tend 
to produce false positive conclusions should be used with a very carefully 
manner in a practical scenario. 
\item Over the five data sets, our proposed $5\times 2$ BCV McNemar's test 
almost achieves the smallest type I errors which are obviously smaller than 
0.05. Therefore, the $5\times 2$ BCV McNemar's test is more conservative than 
the other tests because the bounds of $\rho_1$ and $\rho_2$ in Eq. 
(\ref{eqn:rho_bound}) are loose. Therefore, how to refine the bounds of $\rho_1$
and $\rho_2$ to further improve a McNemar's test is an important future research
direction.
\end{enumerate}  

Figure \ref{fig:power_plots} illustrates the power curves of the tests which own
reasonable type I errors on the experimental data sets.  A common conclusion 
obtained from all the plots in Figure \ref{fig:power_plots} is that the 
$5\times 2$ BCV McNemar's test possesses a more skewed power curve than the 
other tests. The observation indicates that the $5\times 2$ BCV McNemar's test 
is the most powerful one among all the tests for comparing two classification 
algorithms. 

In sum, all the experimental results in this section show that the $5\times 2$ 
BCV McNemar's test possesses not only a reasonably small type I error but 
also a promising power. Thus, the $5\times 2$ BCV McNemar's test is 
superior to the other tests and can reduce false positive conclusions in the
task of algorithm comparison. Therefore, we recommend the $5\times 2$ BCV 
McNemar's test in the practical comparison task of two classification 
algorithms.

\section{Conclusion}
\label{sec:conclusion}
In this study, we proposed a $5\times 2$ BCV McNemar's test for comparing two
classification algorithms. The proposed test is formulated based on a notion 
of ``effective contingency table'' that effectively compress the ten correlated
contingency tables in a $5\times 2$ BCV with a Bayesian perspective. The 
correlations between the tables are well investigated, and 
theoretical bounds of the correlations are proved. Extensive experiments 
illustrate that the $5\times 2$ BCV McNemar's test has a reasonable type I error
and a promising power compared with the existing algorithm comparison methods.

In the future, we will refine the McNemar's test in an $m\times 2$ BCV with a 
sequential setting and further investigate the $m\times 2$ BCV McNemar's test 
from a Bayesian perspective for inferring informative decisions in the task of 
comparing classification algorithms.
  

\ifCLASSOPTIONcompsoc
  \section*{Acknowledgments}
\else
  \section*{Acknowledgment}
\fi

This work was supported by the National Natural Science Foundation
of China under Grants no. 61806115. The experiments are supported
by High Performance Computing System of Shanxi University.

\ifCLASSOPTIONcaptionsoff
  \newpage
\fi



\bibliographystyle{IEEEtran}
\bibliography{cvmcnemar}
%


%





\end{document}